\documentclass[%
 aapm,
 mph,%
 amsmath,amssymb,
 reprint,%
]{revtex4-1}

\usepackage{graphicx}% Include figure files
\usepackage{dcolumn}% Align table columns on decimal point
\usepackage{bm}% bold math

\usepackage{lmodern} 
\usepackage{amsmath}
\usepackage{amssymb}
\usepackage{graphicx,scalerel}
\newcommand\wh[1]{\hstretch{2}{\hat{\hstretch{.5}{#1}}}}

\begin{document}

\preprint{AAPM/123-QED}

%LCLS Analysis Advances: Bi-Cross Validation of the Inverted Laplacian for Cluster Number and Generalized hyper parameter Estimation
%Analysis of X-Ray Scattering Data using Bi-Cross Validation for Estimating hyper parameters and Number of Clusters 
\title[Sample title]{Bi-cross validation for estimating spectral clustering hyper parameters }% Force line breaks with \\
%\thanks{Footnote to title of article.}

\author{Sioan Zohar}
\author{Chun Hong Yoon}
\affiliation{Photon Data and Controls Systems, Linac Coherent Light Source, SLAC National Accelerator Laboratory 2575 Sand Hill Rd, Menlo Park, CA 94025}%Lines break automatically or can be forced with \\
%\author{B. Author}%
%\email{Second.Author@institution.edu.}
%\affiliation{ 
%Authors' institution and/or address%\\This line break forced with \textbackslash\textbackslash
%}%

\date{\today}% It is always \today, today,

\begin{abstract}
One challenge impeding the analysis of terabyte scale x-ray scattering data from the Linac Coherent Light Source LCLS, is determining the number of clusters required for the execution of traditional clustering algorithms. Here we demonstrate that previous work using bi-cross validation (BCV) to determine the number of singular vectors directly maps to the spectral clustering problem of estimating both the number of clusters and hyper parameter values. These results indicate that the process of estimating the number of clusters should not be divorced from the process of estimating other hyper parameters. Applying this method to LCLS x-ray scattering data enables the identification of dropped shots without manually setting boundaries on detector fluence and provides a path towards identifying rare and anomalous events.
\end{abstract}

\keywords{Suggested keywords}%Use showkeys class option if keyword
\maketitle

\section{Introduction}
X-Ray Free Electron Lasers (X-FELs) \cite{ishikawa2012compact} are remarkable instruments capable of producing highly coherent x-ray pulses less than 20 fs in duration.  Since their inception, X-FELs have made contributions to a diverse range of disciplines spanning from condensed matter \cite{higley2019ultrafast} and atomic molecular optics \cite{yang2018imaging} to structural biology \cite{nogly2018retinal} and femto-second chemistry \cite{hong2015element}.
Compared to 3rd generation light sources, X-FELs require high throughput data systems \cite{thayer2016data} for writing to disk on a per-pulse basis. Originally developed in order to filter out low fluence shots in post processing, shot-by-shot recording has since shifted the data collection paradigm and provided researchers with the means to compensate x-ray/laser timing timing jitter \cite{droste2019high}, out run x-ray damage accumulation in protein crystallography experiments \cite{spence2017outrunning,kupitz2017structural}, and offers the potential to extract new physics by identifying rare events \cite{schoenlein2017linac}.

Data accumulated over the course of an LCLS user experiment regularly exceeds 20 TB and approximately 2.5 years analyzing such data is required before the results are published. Efforts to expedite the analysis have motivated the development of a high performance computing infrastructure, user friendly abstraction layers \cite{damiani2016linac}, and novel algorithms \cite{yoon2011unsupervised}. One promising avenue for streamlining data analysis is the exploitation of clustering algorithms. Such algorithms are currently used to cluster diffraction images of protein conformations collected in diffract and destroy experiments \cite{yoon2011unsupervised}, and also have the potential to identify rare events \cite{schoenlein2017linac}. One impediment to this approach is the problem of estimating the hyper parameters and number of clusters required for executing clustering algorithms. Early work estimating the number of clusters used a combination of gap methods \cite{tibshirani2001estimating}, distortion methods methods \cite{sugar2003finding}, stability approaches \cite{tibshirani2005cluster,von2010clustering}, and non parametric methods \cite{fujita2014non}. These approaches are generally considered to be heuristic with well understood limitations and require assumptions about the cluster distribution. More recent work  \cite{fu2017estimating} has made exciting progress in both implementing and laying the theoretical foundation for abstracting Bi-Cross Validation (BCV) \cite{owen2009bi} away from its matrix formulation to estimate the number of clusters for use with the k-means algorithm. This approach \cite{fu2017estimating}, however, requires pre-conditioning rotations to discriminate when multiple clusters are spaced along a single feature dimension and can only label clusters that are linearly separable.  In that work \cite{fu2017estimating}, it was predicted that applying BCV to the Laplacian matrix after the eigen-vector transformation would provide a convex loss function for estimating the number of clusters.

Here, it is shown that spectral clustering hyper parameters, including the number of clusters, can be estimated by performing BCV on the inverted Laplacian matrix and finding the local minima of the resultant BCV loss function. In spectral clustering, data are embedded into a higher dimensional graph representation called the Laplacian matrix  \cite{von2007tutorial,von2010clustering}.  The multiplicity of the Laplacian's smallest eigenvalues is equal to the number of clusters. BCV is a powerful least squares method for estimating the number of dominant singular vectors needed reconstruct the matrix without over fitting the data to the noise \cite{owen2009bi}.  Inverting the Laplacian matrix converts the problem of cluster number estimation from one of estimating the number of smallest singular vectors into the problem of estimating the number of largest singular vectors that, in turn, can be solved using BCV.

The main result of this work is captured in equation \ref{eq:main} which connects the spectral clustering and BCV frameworks. The range where this techniques succeeds and fails is explored using simulated data sets.  Applying this technique to experimental LCLS x-ray scattering data separates low fluence from high fluence x-ray pulses, and provides a path towards identifying clusters of rare events.

\section{Theory}

We consider a set of x-ray scattering data stored within a matrix $\mathbf{X}$,  with elements $\mathbf{X_{i,j}}$ where $\mathbf{i}$ and $\mathbf{j}$ are the rows and columns indices respectively. All entries contained within a row have been measured at the same instant, and all entries within a single column measure the same quantity. For the case of LCLS data, potential column labels are incident x-ray pulse energy, scattered pulse energy, photon energy, x-ray/laser jitter correction, or laser delay stage position. The process of clustering, in this context, means creating columns that assign labels, such as ``signal of interest'', ``low fluence shots'' , ``outliers'', or ``rare events'' to each of the rows.

In the spectral clustering approach, clusters are identified by applying k-means clustering on the $k$ smallest eigen-vectors, $v$, of the Laplacian matrix, $\mathbf{L}$, where $k$ is the number of clusters. Formally, 

\begin{equation}
\mathbf{L} = \mathbf{D} - \mathbf{W}
\end{equation}

where $\mathbf{D}$ is the degree matrix. The weighting $\mathbf{W}$ matrix chosen here is calculated using the radial basis function (RBF) kernel\cite{chung2003radius} such that

\begin{align}
%\mathbf{W_{i,j}} = \mathbf{W_{j,i}} = \exp\left[{-\Gamma\sum_m(\mathbf{X_{i,m}-X_{j,m}})^2}\right]
\nonumber & \mathbf{W_{i,j}} = \mathbf{W_{j,i}} = \\  
& \exp\left[{-\sum_m(\mathbf{X_{i,m}-X_{j,m}})^\mathbf{T}\mathbf{\Gamma}(\mathbf{X_{i,m}-X_{j,m}})}\right]
\end{align}

where $\mathbf{i}$ and $\mathbf{j}$ are the row and column indices of $\mathbf{W}$, and $\Gamma$ is a hyper parameter that is inversely proportional to root of the expected distance between points within a cluster. Traditionally, $\Gamma$ is treated as a scalar. In practice, the Laplacian is normalized by

\begin{equation}
\mathbf{L_n} = \mathbf{D^{-1/2}} \mathbf{L}\mathbf{D^{-1/2}} = \mathbf{I}-\mathbf{D^{-1/2}} \mathbf{W}\mathbf{D^{-1/2}}
\end{equation}

where $\mathbf{L_n}$ is the normalized Laplacian. Using these definitions, the spectral clustering method proceeds by solving the generalized eigen-vector problem

\begin{equation}
\mathbf{L_n}v = \lambda \mathbf{D} v,
\end{equation}

implementing k-means on the diagonalized feature space, and propagating the resultant labels from k-means back to $\mathbf{X}$. 

\begin{figure}
\includegraphics[width=0.5\textwidth]{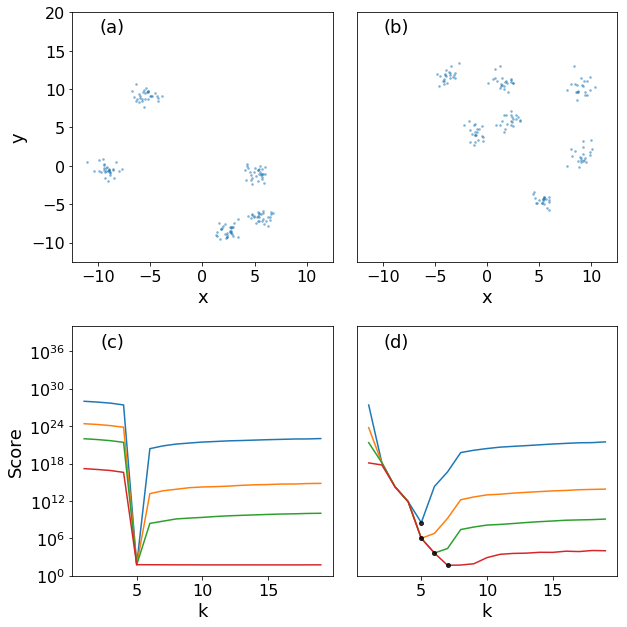}% Here is how to import EPS art
\caption{\label{fig:sample_data} (a) A set of 150 samples occupying a 7D feature space are clustered into 5 groups and projected onto 2D. (b) The inter-cluster spacing is reduced by  reducing the feature space from 7D to 2D and increasing the number of clusters from 5 to 7. (c) The BCV score dependence on the number of clusters for regularization parameter values of 1e-14 (blue), 6.3e-13 (orange), 1e-12(green), 2.5e-09 (red). The score minimum occurs at 5 which is the excepted number of clusters. (d) When the inter-clustering spacing is reduced, BCV does not robustly estimate the number of clusters, since the score minimum (black dots) does not occur at the same $k$ value for all values of $\xi$ and only occurs at the expected value of 7 for $\xi$ = 2.5e-9.}
\end{figure}

The procedure for estimating the number of clusters and $\Gamma$ by performing BCV on the inverted $\mathbf{L}^{-1}$ proceeds as follows.  The Laplacian is, by construction, a singular matrix that cannot be inverted. This drawback is circumvented by add a regularization term, $\mathbf{R}$. That is 

\begin{equation}
\mathbf{L_r} = \mathbf{L_n}+ \xi \mathbf{R}
\end{equation}

where $\xi$ is a scalar regularization parameter.  Here, $\xi$ is empirically determined to be of the order 1e-9 to 1e-14.  The matrix $\mathbf{R}$ is 

\begin{equation}
\mathbf{R} = \mathbf{H} - \mathbf{H^T} \mathbf{L_n}\mathbf{H}
\end{equation}

where $\mathbf{H}$ is a Haar distributed random matrix \cite{mezzadri2006generate}. Adding $\xi\mathbf{R}$ to $\mathbf{L_n}$, as opposed to adding $\xi\mathbf{H}$ directly, guarantees the resultant matrix $\mathbf{L_r}$ can be inverted. The BCV loss function for $\mathbf{L_r}^{-1}$ is calculated as described in \cite{owen2009bi} by breaking $\mathbf{L_r}^{-1}$ into quadrants.

\begin{equation}
\mathbf{L_r}^{-1} = \left[ {\begin{array}{cc}
   \mathbf{A} & \mathbf{B} \\
   \mathbf{C} & \mathbf{E} \\
  \end{array} } \right]
\label{eq:main}
\end{equation}

The bottom right quadrant has been labeled in this work $\mathbf{E}$, deviating from the notation in previous literature \cite{owen2009bi} so as not to be confused with the degree matrix $\mathbf{D}$. Here, $\mathbf{A}$ was designated as the hold out and 2x2 BCV was configured such that the sub matrices A,B,C,D have the same number of rows and columns. This sub-matrix partitioning is close to the optimal 52\%  holdout size for square matrices \cite{perry2009cross}. The BCV loss function is 

\begin{equation}
BCV(k,\Gamma) = \sum_{\mathbf{i,j}} (\mathbf{A} - \mathbf{B} (\wh{\mathbf{E}}^{(k)})^+ \mathbf{C})^2_{\mathbf{i,j}}
\end{equation}

where $(\wh{\mathbf{E}}^{(k)})^+$ is the Penrose pseudo inverse of $(\wh{\mathbf{E}}^{(k)})$

\begin{equation}
(\wh{\mathbf{E}}^{(k)})^+ = ((\wh{\mathbf{E}}^{(k)})^T\wh{\mathbf{E}}^{(k)})^{-1} (\wh{\mathbf{E}}^{(k)})^T
\end{equation}

and $(\wh{\mathbf{E}}^{(k)})$ is the SVD reconstruction of $\mathbf{E}$ using $k$ number of basis vectors. The procedure starting from equation \ref{eq:main} was iterated $\sim$ 40 times with $\mathbf{L_r}^{-1}$ being shuffled each iteration before being decomposed into sub matrices.  The BCV score used to determine the number of clusters is the average BCV score over all iterations.

\begin{figure}
\includegraphics[width=0.5\textwidth]{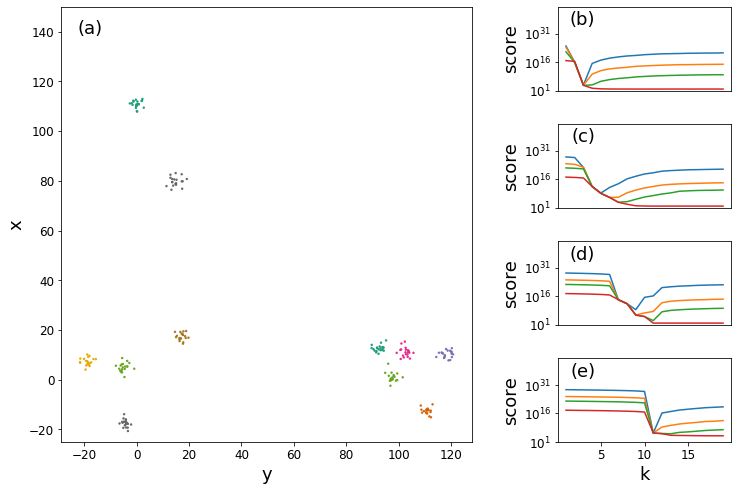}% Here is how to import EPS art
\caption{\label{fig:sigma_dependence} Demonstration of cluster identification at different length scales. (a) a set of 150 samples clustered in to 11 groups that appears as 3 clusters on longer length scales. (b) density map of their score dependence on $\Gamma$ and k. Regularization values are 1e-14, (blue), 6.3e-13(orange), 1e-12(green), 2.5e-09(red). The score as  function of the cluster number $k$ is shown for $\Gamma$ equal to 0.005, 0.028, 0.158 and 1.58 for panels (b), (c), (d), and (e) respectively.}
\end{figure}

\section{Numerical Simulations}

The performance of this approach was bench marked for a range of hyper parameters using scikit-learn version 0.19.1, numpy version 1.14.2, and scipy version 0.19.1 packages \cite{scikit-learn,oliphant2006guide,oliphant2007python,van2011numpy}.  Source code containing an executable step by step walk through can be cloned from this repository \cite{clustering_bcv_step_by_step}.

\begin{figure}
\includegraphics[width=0.30\textwidth]{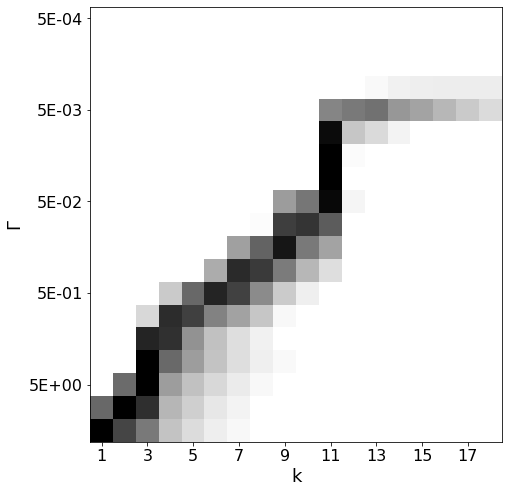}% Here is how to import EPS art
\caption{\label{fig:heatmap} Density map of the score dependence on $\Gamma$ and $k$ for $\xi$ = 1e-14. The dark and light regions correspond to low and high BCV loss function values respectively.  Cross sections of this density map for fixed values of $\Gamma$  are shown in Figure \ref{fig:sigma_dependence} panels (b) through (e).}
\end{figure}

In Figure \ref{fig:sample_data} (a) a set of 5 simulated clusters projected from 7 dimension feature space onto 2 dimensions is shown. Panel (b) shows 7 simulated clusters generated in a 2D feature space. The BCV loss function minimum was found by iterating over increasing values of cluster number, $k$, and length scales, $\Gamma$ and calculating the BCV at loss function each point. The BCV score's dependence on $k$ for the clusters in panel (a) and (b) are shown in panels (c) and (d) respectively. The different color lines shown in (c) and (d) correspond to increasing values of the regularization parameter. The BCV score in (c) has a minimum at $k$ =5 correctly identifying the number of clusters.  This estimate is robust for changing regularization values except for large regularization, where the score minimum no longer occurs at the expected number of clusters and moves to arbitrarily large k. The inter-cluster distance for points in Fig. \ref{fig:sample_data} panel (b) is decreased with respect to panel (a) by increasing the number of clusters from 5 to 7 and reducing the feature space dimension from 7 to 2.  The BCV score for the points in panel (b) are shown in panel (d).  For a fixed value of $\Gamma $, the cluster number estimation procedure is not robust since the score minimum does not reliably estimate the number clusters for all values of $\xi$.

In Figure \ref{fig:sigma_dependence} (a), a set of clusters in 2D are shown.  The clusters can be partitioned into 3 or 11 different groups, depending on the Gaussian kernel width, $\Gamma$, chosen to construct the affinity matrix. The BCV scores plotted as a function of the number of clusters are shown in panels (b), (c), (d), and (e) for values of $\Gamma$ equal to 0.005, 0.028, 0.158 and 1.58  respectively.  The different colored curves are for different values of the regularization parameter $\xi$.  For the smallest regularization values (blue cuves), two global minimum occurring at $\Gamma$ values of 1.58 (Fig.  \ref{fig:sigma_dependence} (b)) and 0.005 (Fig.  \ref{fig:sigma_dependence} (e) ) occur at $k$ equal to 3 and 11 respectively. In Figure \ref{fig:heatmap} a heat map of the BCV\textsc{\char13}s score value's dependence upon the gaussian kernel width and number of cluster is shown for  $\xi = 10^{-14}$.  The RBF parameter $\Gamma$ can be converted into a characteristic length scale $\sigma$, using $\Gamma = 1/(2 \sigma^2)$. The two local minimum observed observed at $k$ = 11 and $k$ = 3 have corresponding $\sigma$ values of the orders of 1 and 10 respectively, which correspond to two different length scales at which the clusters can be grouped. The ability to estimate both the number of clusters and the spectral clustering $\Gamma$ hyper parameter is advantageous compared to previous methods which provide a loss function that estimates the number of clusters but not any additional hyper parameters.

\section{Experimental Data}

%GMD/relativeEnergyPerPulse',
%'delayStage',
%'gas_detector/f_12_ENRC',
%'gas_detector/f_22_ENRC',
%'gas_detector/f_11_ENRC',
%'gas_detector/f_21_ENRC',
%'ebeam/L3_energy',
%'laser_power',
%'recalculated_atm',                    
%'recalculated_atm_rms',
%'scattered_svd_estimated',
%'ebeam/photon_energy'

\begin{figure}
\includegraphics[width=0.35\textwidth]{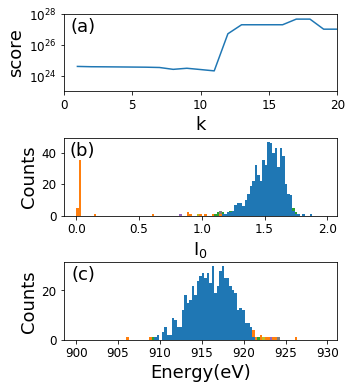}% Here is how to import EPS art
\caption{\label{fig:real_data} (a) the BCV score minimum occurs for 11 clusters. (b) histogram the incident pulse energy measured in a gas detector upstream the monochromator.  The orange and blue histogram the dropped shots and signal of interest respectively. (c) histogram of the photon energy generated upstream the monochromator. }
\end{figure}

In Figure \ref{fig:real_data} the results from applying this approach to  x-ray scattering experiment are shown. The feature space is 12 dimensions with column labels corresponding to intensity of x-rays scattered off the sample, incident intensity downstream the monochromator, 4 different incident intensity diagnostics from upstream the monochromator,  laser delay stage position, laser  power, arrival time monitor mean and FWHM, photon energy, and the photon energy product with the incident intensity down stream the monochromator.  Multiplying the photon-energy with the intensity linearizes the chromatic non-linearity observed when the photon energy is tuned to the steep part of an x-ray absorption edge \cite{zohar2019multivariate}. 

The problem of heterogeneous density present in spectral clustering is circumvented by feature engineering an additional column that contains an estimate of the point density in the local vicinity.  This was accomplished by appending the diagonal values of the degree matrix, calculated for 7000 samples using a $\Gamma$ = 1e-2, to the feature space. Clustering was performed on a total of 750 rows from this feature space. Eleven clusters are identified with the populations of the dominant first three clusters containing on average 573, 62, 43 data points. The rest of the data points are spread over the remaining clusters. As shown in Figure \ref{fig:real_data} this approach separates out the dominant cluster (blue histogram) which corresponds to signal of interest from the dropped shots with no fluence (orange histogram). It is stressed that the last figure presented here is analyzed on less than 1 \% of the entire data and does not represent the expected number of clusters if the full data set were to be used.

\section{Discussion}

There are several advantages for using the matrix formulation \cite{owen2009bi} of BCV as opposed to the abstracted BCV form in the non embedded feature space \cite{fu2017estimating}.  One advantage is that the pre-conditioning rotation steps needed for preventing clusters from laying along one non-separable dimension are no longer required.  Another advantage is that since the matrix BCV formulation does not require a classification step, there are no additional hyper-parameters that need to be estimated.

The ability to simultaneously estimate both the $\Gamma$ parameter and number of clusters is not serendipitous.  Intuitively, it is readily understood that asking ``how many clusters are present in some region'' can not be separated from the question of``what length scales do those same clusters appear on?''  This line of thinking agrees with the limiting cases  of very small and very large $\Gamma$ values, where the number of estimated clusters will be equal to either one or the number of points respectively. Looking forward, there are several pre-requisites that would need to be met for this approach to be widely adopted.  A mathematical proof demonstrating that the BCV loss function minimum correctly estimates the hyper parameters would have to be shown. This proof would provide insight on how to estimate the regularization parameter by exploiting the regularized Laplacian's condition number and using eigen-vector decomposition of the inverted Laplacian as opposed to SVD decomposition. 

%The spectral clustering algorithm used here employed the RBF kernel to convert the adjacency matrix $L^2$ into a weighting matrix.  This particular RBF kernel has a single scalar parameter, $\Gamma$. The approach demonstrated here to optimize $\Gamma$ could be used to generalize $\Gamma$ from a scalar to a covariance matrix (or more complicated kernel)  with each matrix element being optimized using the BCV approach. Scanning such a hyper parameter space would require $O(m^n)$ time, where $n$ is the number of hyper parameters and $m$ is the number of points. A more efficient minimum finding approach exploiting gradient descent methods would be required to reduce the time complexity. Initializing gradient descent methods at different hyper parameter values would provide a path towards avoiding the trivial solution describing a single cluster at large length scales.

\section{Conclusion}

In conclusion, a direct matrix implementation of BCV for estimating both the number of clusters and kernel hyper parameters used in spectral clustering has been demonstrated.  This was accomplished by applying the matrix formulation for BCV directly to the inverted Laplacian matrix. The resulting BCV loss function has robust minima that occur at different cluster numbers depending upon the length scales determined by RBF kernel parameter. The results here provide a path towards generalized hyper parameter optimization for spectral clustering algorithms.

We thank Art B. Owen for providing fruitful discussions and insights. This work was performed in support of the LCLS project at SLAC supported by the U.S. Department of Energy, Office of Science, Office of Basic Energy Sciences, under Contract No. DE-AC02-76SF00515.

%\nocite{*}

\bibliography{aapmsamp}% Produces the bibliography via BibTeX.

\end{document}